\documentclass[letterpaper, 10 pt, conference]{ieeeconf}
\IEEEoverridecommandlockouts
\usepackage{amsmath,amssymb,amsfonts}
\usepackage{algorithmic}
\usepackage{graphicx}
\usepackage{textcomp}
\usepackage{graphbox}
\usepackage{tikz,tikz-3dplot}
\usepackage{pgfplots}
\usetikzlibrary{pgfplots.groupplots}
\usetikzlibrary{arrows,arrows.meta,automata,backgrounds,calc,chains,%
decorations.markings,decorations.pathreplacing,decorations.pathmorphing,%
matrix,positioning,shapes,shapes.geometric,shapes.symbols,spy,trees,tikzmark}
\usepackage[product-units=single,per-mode=symbol,range-units=single,detect-all]{siunitx}
\DeclareSIUnit\pixel{px}
\usepackage{xcolor}
\usepackage{booktabs}
\usepackage[bibencoding=utf8,sorting=none,doi=false,maxbibnames=14,minbibnames=14]{biblatex}
\usepackage{balance}

\usepackage[nolist]{acronym}
\begin{acronym}
\acro{UAV}{unmanned aerial vehicle}
\acro{UGV}{unmanned ground vehicle}
\acro{GCS}{ground control station}
\acro{GNSS}{global navigation satellite system}
\acro{MC}[WMC]{water monitor controller}
\end{acronym}

\newcommand{\wrt}{w.r.t.~}

\newcommand{\eg}{e.g.,\ }
\newcommand{\cf}{cf.,\ }
\newcommand{\etal}{~et al.~}
\newcommand{\reffig}[1]{Fig.~\ref{#1}}
\newcommand{\refsec}[1]{Sec.~\ref{#1}}
\newcommand{\reftab}[1]{Tab.~\ref{#1}}

\def\WithAuthorInfo{1}

\def\BibTeX{{\rm B\kern-.05em{\sc i\kern-.025em b}\kern-.08em
    T\kern-.1667em\lower.7ex\hbox{E}\kern-.125emX}}

\title{\LARGE \bf
Aerial Assistance System for Automated Firefighting during\\Turntable Ladder Operations
}
\if\WithAuthorInfo1
\author{Jan Quenzel~${}^{*,a,b}$, Valerij Sekin~${}^{*,c}$, Daniel Schleich~${}^{a,b}$, Alexander Miller~${}^c$, \\ Merlin Stampa~${}^{c,d}$, Norbert Pahlke~${}^e$, Christof Röhrig~${}^c$ and Sven Behnke~${}^{a,b}$
\thanks{a)~Autonomous Intelligent Systems Group, Computer Science Institute VI -- Intelligent Systems and Robotics; ~b)~Center for Robotics and Lamarr Institute for Machine Learning and Artificial Intelligence, University of Bonn, Germany;
~c)~IDiAL -- University of Applied Sciences and Arts Dortmund, Germany; 
~d)~German Rescue Robotics Center (DRZ), Germany;
~e)~Institute for Fire Service and Rescue Technology -- Fire Department of Dortmund, Germany; ~*)~denotes equal contribution;
{\tt\small quenzel@ais.uni-bonn.de}}
}
\else
\author{Anonymous Authors
}
\fi
\begin{document}
\maketitle
\thispagestyle{empty}
\pagestyle{empty}
\begin{tikzpicture}[remember picture,overlay]
  \node[anchor=north,align=center,font=\sffamily\small,yshift=-0.4cm] at (current page.north) {%
  \textbf{Accepted final version.} IEEE International Symposium on Safety, Security, and Rescue Robotics (SSRR), Galway, Ireland, October 2025
  };
\end{tikzpicture}%

\begin{abstract}
Fires in industrial facilities pose special challenges to firefighters, \eg due to the sheer size and scale of the buildings.
The resulting visual obstructions impair firefighting accuracy, further compounded by inaccurate assessments of the fire's location.
Such imprecision simultaneously increases the overall damage and prolongs the fire-brigades operation unnecessarily.

We propose an automated assistance system for firefighting using a motorized fire monitor on a turntable ladder with aerial support from an \ac{UAV}.
The \ac{UAV} flies autonomously within an obstacle-free flight funnel derived from geodata, detecting and localizing heat sources.
An operator supervises the operation on a handheld controller and selects a fire target in reach.
After the selection, the \ac{UAV} automatically plans and traverses between two triangulation poses for continued fire localization.
Simultaneously, our system steers the fire monitor to ensure the water jet reaches the detected heat source.
In preliminary tests, our assistance system successfully localized multiple heat sources and directed a water jet towards the fires.
\end{abstract}

\section{Introduction}
Fires in high-rise or industrial buildings are uniquely challenging and dangerous for firefighters and residents alike~\cite{wang2025HighRiseSurvey}.
In these situations, proximity to fire and smoke is not the only problem.
Debris, the danger of collapse, and potentially hazardous materials~\cite{vallee2021LeadNotreDame} pose risks to the lives of first responders.
Moreover, assessing the fire's position is often difficult for firefighters~\cite{GrenFellFireReport}.
Misjudgments lead to inefficient extinguishing, higher water consumption, delays in rescuing people, and endangering emergency services.
Visual obstructions, due to building size, other buildings, narrow streets, smoke or trees~\cite{roldanGomez2021survey}, further hinder fire-fighting operations.
Maintaining an overview becomes increasingly difficult and strenuous, as experienced in the tragic Grenfell Tower fire~\cite{GrenFellFireReport}.
As a consequence, the London fire brigade procured \acp{UAV} to improve their situational awareness during deployments to residential high-rise buildings.

Coordinating the fire extinguishing from the ladder based on images captured by a UAV remains challenging, though.
To overcome this issue, we propose an aerial assistance system for semi-autonomous firefighting during turntable ladder operations.
We combine an autonomous \ac{UAV} and an automated fire monitor for semi-autonomous extinguishing.
The commercial \acs{UAV} starts autonomously in a collision-free flight funnel computed from 3D geodata~\cite{quenzel2024gnss} and \acs{GNSS}.
Our \textit{ground control station} (GCS) processes the live imagery from a thermal and a color camera to detect and localize heat sources in a georeferenced frame.
A \textit{handheld controller} (HHC) provides visual feedback to the firefighters including fire location, nozzle orientation, and water jet prediction.
The firefighter selects one of the detected fires as target and supervises the subsequent extinguishing. 
The georeferenced fire location is sent to our \textit{water monitor controller} (WMC) and the \acs{UAV} alternates between automatically generated observation poses to continuously localize the heat source.
The WMC converts the fire water monitor's \acs{GNSS} pose and the target location to compute the needed nozzle orientation using a ballistic water jet prediction model.

\begin{figure}[t!]
\centering
\includegraphics[trim=300 0 25 0,clip,width=\linewidth]{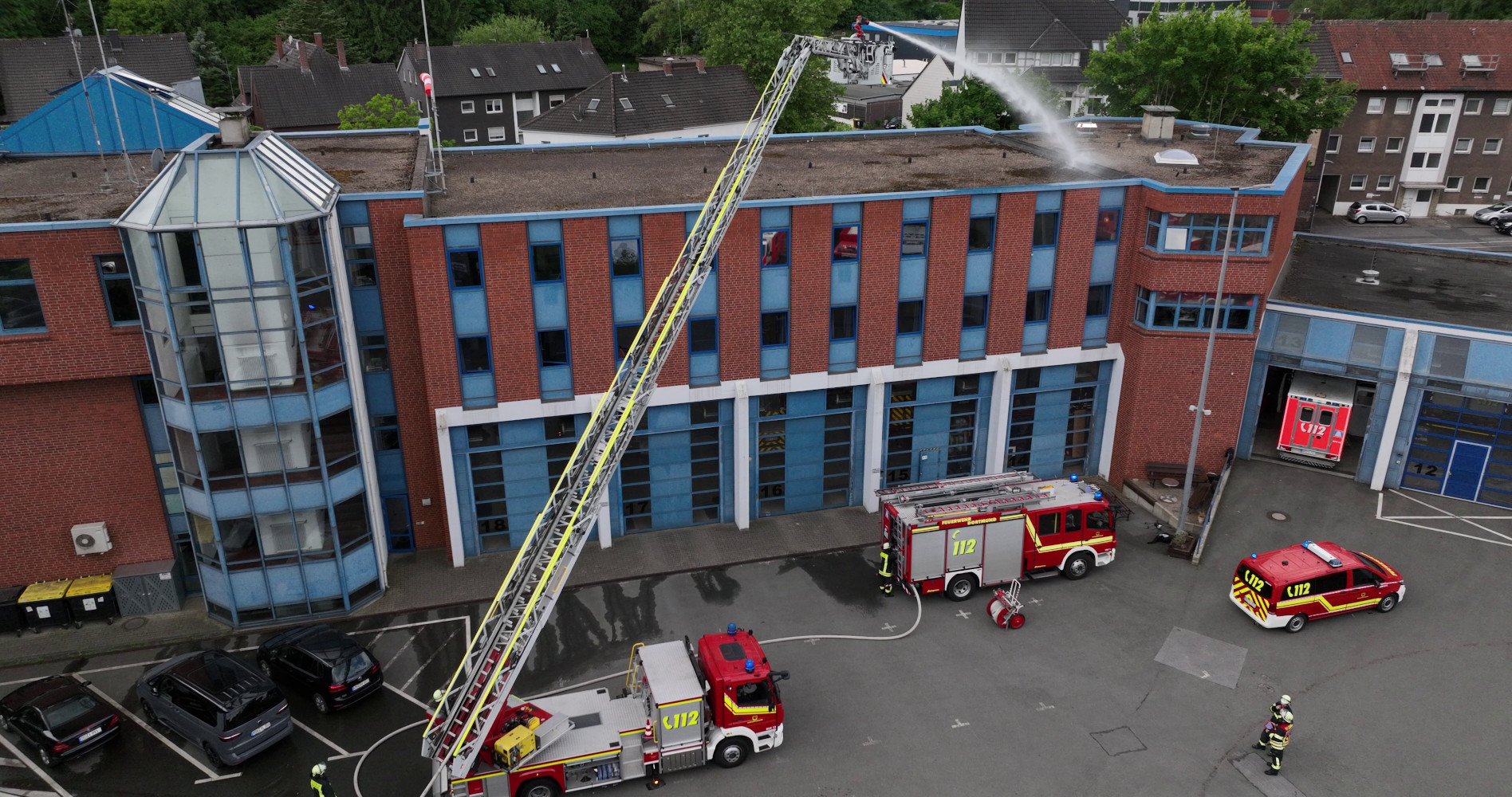}
\caption{An automated fire monitor sprays water from an aerial turntable ladder onto a simulated heat source as detected by the \ac{UAV}.}
\label{fig:teaser}
\end{figure}

Our system facilitates operations without endangering firefighters in the aerial ladder's cage, thus, reducing stress and harm for the operator.
Moreover, the modularity of our approach allows us to mount the fire monitor on a \ac{UGV} for access to remote unsafe areas.

\noindent In particular, our system includes:
\begin{itemize}
    \item an \ac{UAV} flying autonomously in an obstacle-free funnel, detecting and localizing heat sources,
    \item an automated fire monitor for directed water jet application during fire-fighting operations,  and
    \item an intuitive user interface on a handheld controller for system configuration and supervision.
\end{itemize}

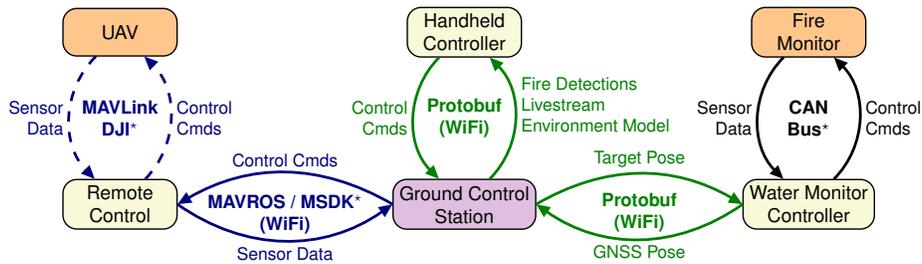
\begin{figure*}
  \centering
  \resizebox{0.7\linewidth}{!}{\begin{tikzpicture}
[content_node/.append style={font=\sffamily,minimum size=2.5em,minimum width=6em,draw,align=center,rounded corners,scale=0.55},
label_node/.append style={font=\sffamily,scale=0.5},
group_node/.append style={font=\sffamily,dotted,align=center,rounded corners,inner sep=1em,thick},>={Stealth[inset=0pt,length=4pt,angle'=45]}]

\definecolor{yellow}{rgb}  {0.85,0.85,0.0}
\definecolor{red}{rgb}     {0.5,0.0,0.0}
\definecolor{green}{rgb}   {0.0,0.5,0.0}
\definecolor{blue}{rgb}    {0.0,0.0,0.5}
\definecolor{grey}{rgb}    {0.5,0.5,0.5}

\node(UAV)[content_node,fill=orange!45!white] at (1.0,2.25) {UAV};
\node(RC)[content_node,fill=yellow!15!white] at (1.0,0.5) {Remote\\Control};
\node(GCS)[content_node,fill=violet!25!white] at (4.5,0.5) {Ground Control\\Station};
\node(MC)[content_node,fill=yellow!15!white] at (8.0,0.5) {Water Monitor\\Controller};
\node(HC)[content_node,fill=yellow!15!white] at (4.5,2.25) {Handheld\\Controller};
\node(Monitor)[content_node,fill=orange!45!white] at (8.0,2.25) {Fire\\Monitor};

\draw[->, thick, dashed, bend right=45, blue] (UAV) to node[label_node,midway,left,align=right] {Sensor\\Data} (RC);
\draw[->, thick, dashed, bend right=45, blue] (RC) to node[label_node,midway,right,align=left] {Control\\Cmds} (UAV);
\draw[draw=none, blue] (RC) to node[label_node,midway,align=center,scale=1.05] {\textbf{MAVLink}\\\textbf{DJI${}^\star$}} (UAV);

\draw[->, thick, bend right=35, blue] (RC.0) to node[label_node,midway,below,align=center] {Sensor Data} node[label_node,midway,above,align=center,scale=1.05] {\textbf{MAVROS / MSDK${}^\star$}\\\textbf{(WiFi)}} (GCS.180);
\draw[->, thick, bend right=30,blue] (GCS.180) to node[label_node,midway,above,align=center] {Control Cmds}(RC.0);

\draw[->, thick, bend left=30, green] (GCS.0) to node[label_node,midway,above,align=left] {Target Pose} (MC.180);
\draw[->, thick, bend left=35, green] (MC.180) to node[label_node,midway,above,align=center,scale=1.05] {\textbf{Protobuf}\\\textbf{(WiFi)}} node[label_node,midway,below,align=center] {GNSS Pose} (GCS.0);

\draw[->, thick,bend right=45] (MC) to node[label_node,midway,right,align=left] {Control\\Cmds} (Monitor);
\draw[->, thick,bend right=45] (Monitor) to node[label_node,midway,left,align=right] {Sensor\\Data} (MC);
\draw[draw=none] (Monitor) to node[label_node,midway,align=center,scale=1.05] {\textbf{CAN}\\\textbf{Bus${}^\star$}} (MC);

\draw[<-, thick, bend left=45, green] (HC) to node[label_node,midway,right,align=left,yshift=0.8em] {Fire Detections\\Livestream\\Environment Model} (GCS);
\draw[<-, thick, bend left=45, green] (GCS) to node[label_node,midway,left,align=right] {Control\\Cmds}(HC);
\draw[draw=none,green] (GCS) to node[label_node,midway,align=center,scale=1.05] {\textbf{Protobuf}\\\textbf{(WiFi)}} (HC);

\end{tikzpicture}}
  \caption{System components and data flow. 
  \acs{UAV} sensor data is send over the remote control to the ground control station (GCS) where the data is processed to localize fires. An operator selects on the handheld controller (HHC) the fire target. The GCS sends the fire's georeferenced coordinates wirelessly to the water monitor controller (WMC) to orient the nozzle of a fire monitor on top of the ladder.}
  \label{fig:data_flow}
\end{figure*}

\section{Related Work}
\acp{UAV} are increasingly applied in firefighting operations~\cite{Kruijff-Korbayova:SSRR21,lattimer2023use}.
The aerial imagery helps to detect fire~\cite{xu2021forest} or smoke~\cite{feiniu2019smoke, rosu2019heat, zhu2020smokeSeg}.
Lu\etal\cite{lu2022vision} combine such detections with binocular stereo depth estimates and GNSS into a \ac{UAV} inspection system that reports georeferenced wildfire locations.
Rosu\etal\cite{rosu2019heat} build a 3D mesh from LiDAR with texture from a thermal camera.
By utilizing the observations of multiple \acp{UAV}, Sherstjuk\etal\cite{sherstjuk20203d} reconstruct 3D frontiers of forest fires.
A survey of different \ac{UAV} systems for fighting wildfires has been compiled by \textcite{keerthinathan2023exploring}. 

For the robotic competition MBZIRC 2020, several \acp{UAV} have been proposed that demonstrated autonomous search and extinction of small simulated fires~\cite{real2021autonomous,martinez2021skyeye,walter2022extinguishing,quenzel2021icuas,beul2022fr}. Viegas\etal\cite{viegas2022tethered} designed a tethered \ac{UAV} to extinguish real fires.

In real-world operations, \acp{UAV} are used to quickly generate an overview of the environment and improve situational awareness.
In contrast, ground robots can additionally carry fire extinguishing devices.
Kong\etal\cite{kong2024computer} propose such a \ac{UGV} that autonomously patrols construction sites. 
Once a fire is detected within the camera images, an ultrasonic sensor provides distance measurements, and a human operator remotely controls the water monitor to extinguish the fire.

Fully automated fire monitors have also been proposed.
Zhu\etal\cite{zhu2020fire} detect the fire target using a thermal camera which is aligned with the nozzle.
Thus, they can use the detected pixel coordinates as feedback to control the fire monitor's yaw angle.
Adjusting the pitch requires an additional prediction of the water jet's falling position~\cite{zhu2020jet}.
A different approach is realized by McNeil\etal\cite{mcneil2013fire}.
They directly estimate the 3D fire localization using stereo thermal cameras and employ a fixed model of the water jet trajectory for controlling the monitor.
In contrast, Lin\etal\cite{lin2021jet} update the water jet model based on the detected landing point.

While the above methods require a direct line-of-sight from fire monitor to target, our approach employs a \ac{UAV} for fire localization.
As a result, our method overcomes possible visual obstructions.

\section{Automated Fire Extinguishing System}
Our approach combines an automated fire monitor on a turntable ladder, an autonomous \acs{UAV}, and a \ac{GCS} into a semi-autonomous fire extinguishing system. \reffig{fig:data_flow} gives an overview of the data flow between the components of our system.

The \acs{UAV} observes the scene using thermal and color cameras while alternating between two triangulation poses.
Meanwhile, the \ac{GCS} computes a georeferenced 3D model and locates heat sources from the live \acs{UAV} imagery.
A handheld controller (HHC) facilitates intuitive configuration and supervision of the autonomous fire extinguishing.
The operator selects one of the detected heat sources and if in range the fire monitor orients itself such that the water jet reaches the fire.

\subsection{Automated Firefighting Monitor}
\begin{figure}
\centering
\resizebox{0.7\linewidth}{!}{\input{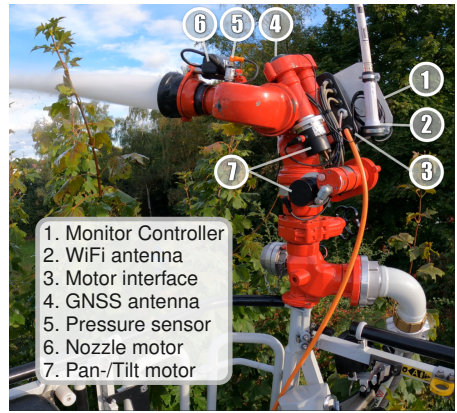}}
\caption{Automated fire monitor on the turntable ladder.}
\label{fig:hardware}
\end{figure}

Our system uses a commercially-available firefighting monitor (\reffig{fig:hardware}) for aerial ladder trucks from Magirus. 
The firefighting monitor has three electrical actuators, one for the adjustable nozzle and two for positioning (pan and tilt).
The nozzle supports two jet modes, one for spray and one for a solid jet.
Each actuator is equipped with electric absolute encoders to determine the position of the axis.
The encoders are read by the monitor's internal control unit which evaluates the angular position.
A pressure sensor, which replaces the standard barometer near the nozzle, is used to reports the pressure of the water jet.

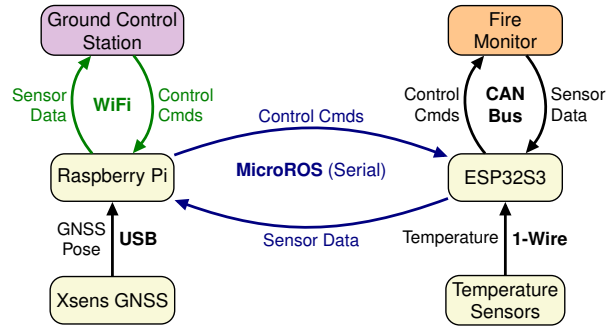
\begin{figure}
  \centering
  \resizebox{0.95\linewidth}{!}{\begin{tikzpicture}
[content_node/.append style={font=\sffamily,minimum size=2.5em,minimum width=6em,draw,align=center,rounded corners,scale=0.55},
label_node/.append style={font=\sffamily,scale=0.5},
group_node/.append style={font=\sffamily,dotted,align=center,rounded corners,inner sep=1em,thick},>={Stealth[inset=0pt,length=4pt,angle'=45]}]

\definecolor{yellow}{rgb}  {0.85,0.85,0.0}
\definecolor{red}{rgb}     {0.5,0.0,0.0}
\definecolor{green}{rgb}   {0.0,0.5,0.0}
\definecolor{blue}{rgb}    {0.0,0.0,0.5}
\definecolor{grey}{rgb}    {0.5,0.5,0.5}

\node(GCS)[content_node,fill=violet!25!white] at (1,2.75) {Ground Control\\Station};
\node(RPI)[content_node,fill=yellow!15!white] at (1,1.25) {Raspberry Pi};
\node(GPS)[content_node,fill=yellow!15!white] at (1.0,0) {Xsens GNSS};
\node(MC)[content_node,fill=yellow!15!white] at (5,1.25) {ESP32S3};
\node(TS)[content_node,fill=yellow!15!white] at (5,0) {Temperature\\Sensors};
\node(Monitor)[content_node,fill=orange!45!white] at (5.0,2.75) {Fire\\Monitor};

\draw[<-, thick, green, bend right=40] (GCS) to node[label_node,midway,left,align=right] {Sensor\\Data} (RPI);
\draw[->, thick, green, bend left=40] (GCS) to node[label_node,midway,right,align=left] {Control\\Cmds} (RPI);
\draw[draw=none, green] (GCS) to node[label_node,midway,align=center,scale=1.05] {\textbf{WiFi}} (RPI);

\draw[->, thick, blue, bend left=20] (RPI) to node[label_node,midway,above,align=center] {Control Cmds} (MC);
\draw[<-, thick, blue, bend right=20] (RPI) to node[label_node,midway,below,align=center] {Sensor Data} 
node[label_node,midway,above,align=center,scale=1.05] {\textbf{MicroROS} (Serial)\\\\} (MC);

\draw[<-, thick] (RPI) -- node[label_node,midway,left,align=center] {GNSS\\Pose} node[label_node,midway,right,align=center,scale=1.05] {\textbf{USB}} (GPS);

\draw[<-, thick] (MC) -- node[label_node,midway,left,align=left] {Temperature} node[label_node,midway,right,scale=1.05] {\textbf{1-Wire}} (TS);

\draw[->, thick, bend left=40] (MC) to node[label_node,midway,left,align=right] {Control\\Cmds} (Monitor);
\draw[<-, thick, bend right=40] (MC) to node[label_node,midway,right,align=left] {Sensor\\Data}  (Monitor);
\draw[draw=none] (MC) to node[label_node,midway,align=center,scale=1.05] {\textbf{CAN}\\\textbf{Bus}} (Monitor);

\end{tikzpicture}}
  \caption{Water monitor controller communication architecture.}
  \label{fig:monitor_controller_architecture}
  \vspace{0.5em}
\end{figure}

The monitor can be controlled in two ways.
One option is the manufacturers control unit, which is typically installed in the aerial ladder basket.
The other option are manually operated hand wheels.
These hand wheels attach to the gearboxes through square couplings.
Using the hand wheels requires the deactivation of the electrical actuators.
As part of the project, the original control unit was moved from the basket to the monitor itself.
The monitor's control unit communicates with other systems on the truck via a CAN-Bus-Interface. We developed the \acf{MC} as an additional control unit to extend the functionality and to integrate the monitor into the overall system, as shown in \reffig{fig:monitor_controller_architecture}.
The \ac{MC} consists of a Raspberry Pi 4 and an ESP32S3 microcontroller.
The Raspberry Pi runs Ubuntu 24.04 with ROS 2 Jazzy.
It acts as the central ROS node for the monitor.
The microcontroller handles the CAN communication with the monitor through a CAN transceiver. 
It connects to the Raspberry Pi via a serial USB-connection and uses Micro-ROS as the protocol.
As shown in \reffig{fig:monitor_controller_architecture}, our \ac{MC} records and processes multiple sensor signals:
\begin{itemize}
\item Temperature measurements inside the \ac{MC} housing and on the exterior of the monitor,
\item Pressure from the monitor's build-in sensor,
\item Position and orientation from an Xsens MTi-680G GNSS module.
\end{itemize}
These sensor values are collected and distributed at a fixed \SI{10}{\hertz}.
The microcontroller acts as a bridge between the monitor's CAN bus and the system's ROS network.
It decodes the incoming ROS messages and translates them into CAN messages for the monitors actuators.
The ESP32 then encodes the status data (position, pressure and temperature) back into ROS messages to publish these into the ROS network.

Our system supports two operating modes:
\begin{itemize}
    \item Manual: The firefighter uses GUI-elements on our user interface to move each actuator manually.
    \item Automatic: A GNSS coded target pose (\eg fire source) is provided to calculate the monitor's target orientation.
\end{itemize}

We use the current nozzle direction, GNSS position and sensor data to model the water jet with a ballistic trajectory.
The \ac{GCS} sends the position of a selected heat source (\cf \refsec{sec:map}) as a ROS message via a local WiFi using Ubiquiti Bullets to the Raspberry Pi.
Based on the relative pose between the monitor's GNSS position and the target, the \ac{MC} calculates the necessary monitor orientation to align the water jet trajectory to the desired target.
If no new target or pose is received, the system maintains the last orientation.
A control loop sends velocity commands to the monitor's actuators to match the desired angular position to orient the nozzle.
These calculations are performed on the Raspberry Pi to provide high flexibility and independence from the central system.
Defined ROS topics enable a manufacturer-independent communication between the \ac{GCS} and the firefighting monitor.

In case of a network failure or other errors during automatic operation, the operator switches to the manual control mode.
This enables the firefighter to still work at a safe distance to the fire.
In case of further disruptions, the mechanical hand wheels can be used as a last fallback method.
As the nozzle does not contain a valve, the operators on the ground remain in control of the water supply. 
This ensures the monitor's operability even in complex real-world scenarios.

\subsection{UAV Mission Control}
\begin{figure}[t!]
\centering
\resizebox{\linewidth}{!}{\includegraphics[trim=0 0 0 0,clip]{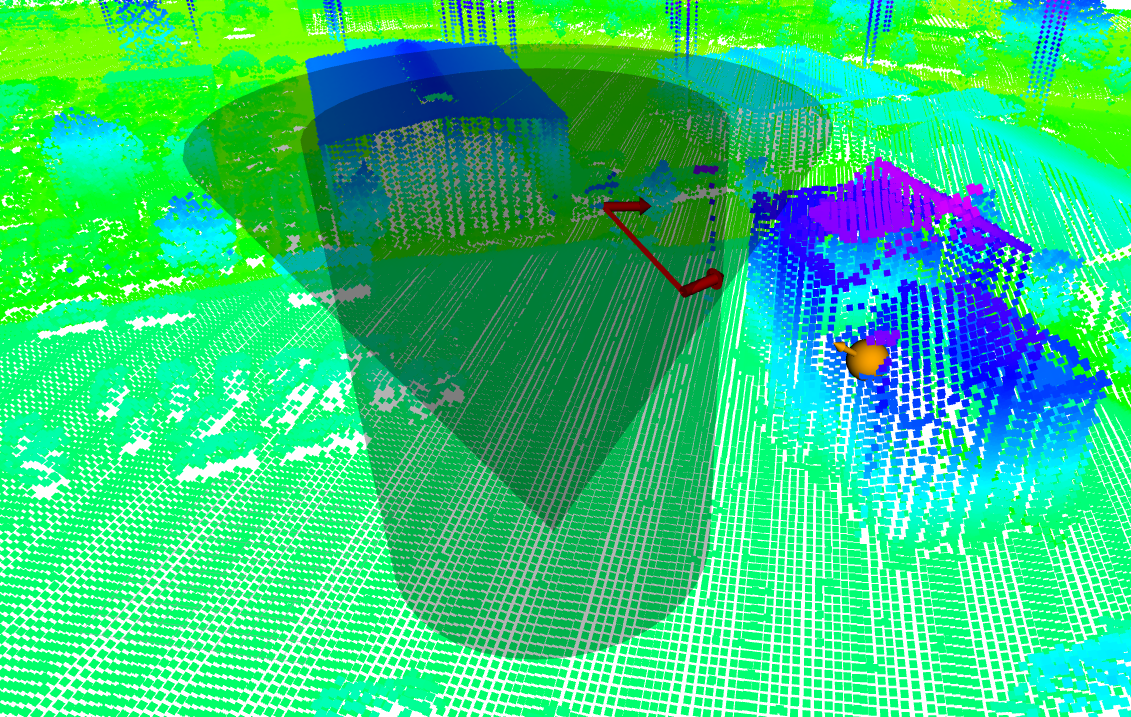}}
\caption{User-defined obstacle-free flight funnel (gray), for safe autonomous operation within a 3D geodata model (green to purple). Two triangulation poses (arrows) are planned relative to the detected heat source (orange).}
\label{fig:funnel}
\end{figure}

To ensure safe autonomous operation, we use an obstacle-free flight funnel within a 3D geodata model~\cite{quenzel2024gnss}, \eg from CityGML or DEM, as shown in \reffig{fig:funnel}.
For this, the funnel center is initialized with the UAV's GNSS pose.
The operator manually adjusts the center within the model, whereas the other parameters are automatically adapted to fit the geodata.
Initially, we turn the 3D model into a 3D occupancy grid and compute a cylinder and cone originating from the center.
For the cylinder, we start with the horizontal distance to the first obstacle and subtract a safety margin to get the cone radius.
The inverse cone describes the visibility from the starting position.
We compute the slope from the ray starting at the origin to the first obstacle.
Our flight funnel is now given by the union of inverse cone and cylinder.

The UAV starts within this funnel using GNSS-based localization. It flies to a predefined height before approaching two initial exploration poses to detect heat sources (\cf \refsec{sec:map}).
Once a fire target is selected, we plan two triangulation poses to maintain visibility of the fire.
We extend the average view direction or surface normal associated with the heat source into the funnel.
Along this ray, we obtain the poses by offsetting to the left and right at a predefined distance within the funnel and orient the gimbal towards the fire.

\subsection{Environment Mapping and Heat Source Localization}\label{sec:map}
Localizing the UAV with GNSS is sufficient for control, yet, too inaccurate for heat sources.
Instead, the GNSS pose and gimbal orientation serve as a prior estimation for an incremental keyframe-based mapping.
The front end extracts Xfeat~\cite{potje2024cvpr} and matches with LightGlue~\cite{lindenberger2023lightglue} while the back end optimizes the observations with GLOMAP~\cite{pan2024eccv}.
We densify the map by triangulating pairwise correspondences from MAST3R~\cite{leroy2024eccv} between keyframes using the previously optimized poses.

To localize heat sources, we adjust the thermal camera settings of our DJI M3T to reduce the amplification of local thermal variations.
Otherwise, a human may appear similarly bright as a fire in the thermal image.
The M3T additionally reports the minimal and maximal temperature with their resp. pixel coordinates within the central image region.
Knowing the pixel corresponding to the highest temperature is a good prior for single heat sources, but insufficient for multiple fires.
Hence, we apply thresholding on the thermal image and extract contours of high intensity regions. 
From these regions, we extract bounding boxes and merge overlapping ones.
In parallel, a new image is retained after traversing a certain distance (\eg \SI{5}{m}) between the planned observation poses.
This triggers our continuous triangulation of the two most recent images using MAST3R and allows us to adapt to a changing environment, \eg due to the fire damage.
Afterwards, we rescale the depth such that the distance between MAST3R's estimated camera poses matches the actual distance traveled \wrt GNSS.
At last, the rescaled depth is projected into all new detections to obtain the source's 3D position.

\subsection{Water Jet Detection}
With the fire localized, we check which detection is in range using the WMC's GNSS and a water jet model.
Once the fire is selected, our system adjusts the direction of the water jet's trajectory.
To detect the water jet, we initially project the origin and direction of the monitor as well as a water jet model into the \ac{UAV}'s images.
This limits the search space within the images.
Thresholding in the temperature range of the water, followed by erosion and dilation, further reduces the candidate area.
We segment the water jet within the candidate area using SAM 2.1~\cite{ravi2025sam2}, thus enabling jet adjustment to cover the heat source with water.

\subsection{User Interface}
\begin{figure}[htb]
    \centering
    \includegraphics[width=\linewidth]{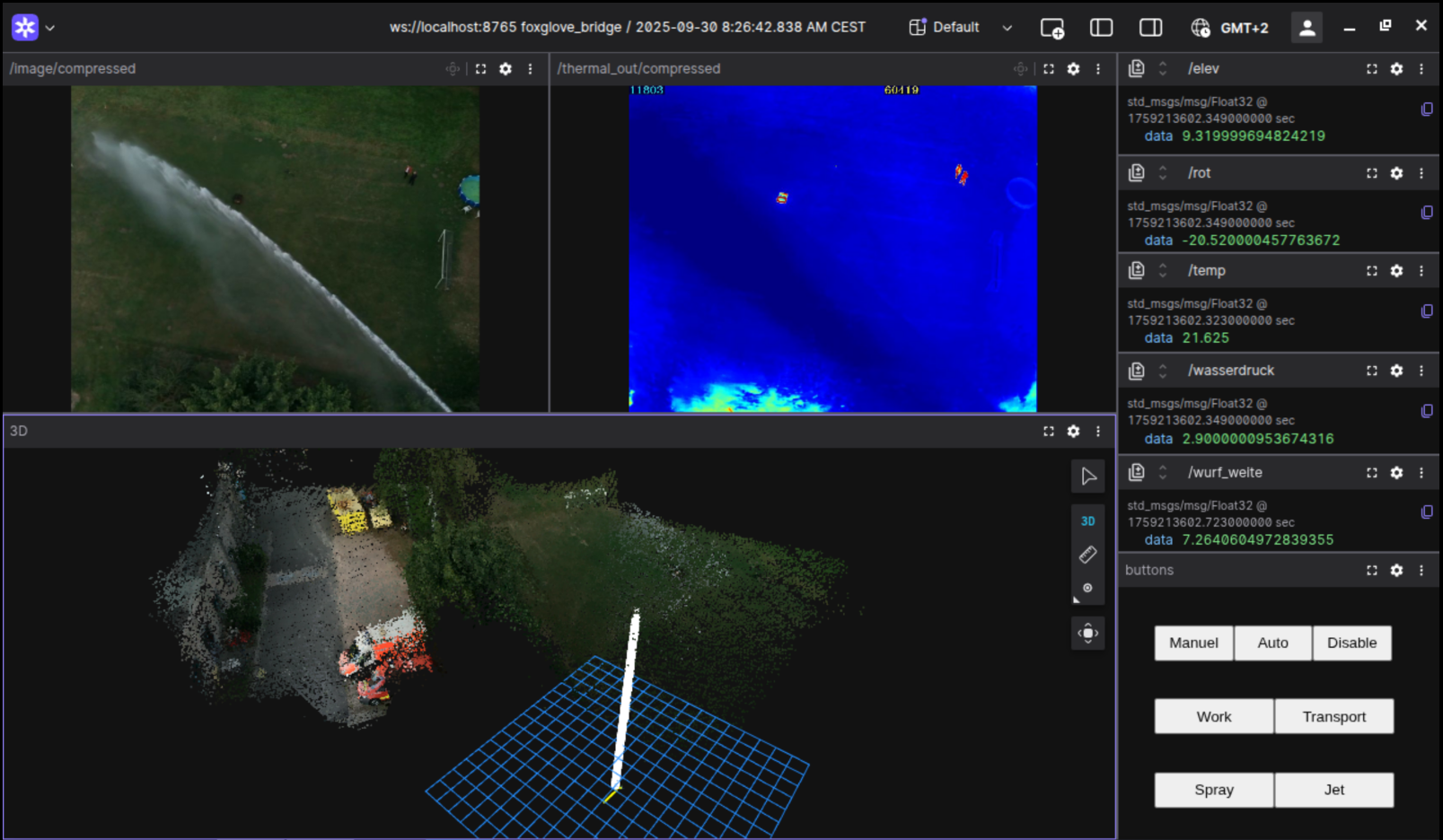}
    \caption{User interface on a handheld Steam Deck controller. The operator configures the system with a control panel (bottom right) and oversees the mission with the UAV's live camera and thermal view (top left, center). The estimated water jet trajectory is visualized within the 3D environment map (bottom).
    }
    \label{fig:UI}
\end{figure}
A Steam Deck\footnote{\url{https://store.steampowered.com/steamdeck/}} allows an operator to supervise and control our system using a compact handheld controller.
Our UI is a custom extension for Foxglove Studio\footnote{\url{https://foxglove.dev/studio}} which is visualized on the Steam Deck, as shown in \reffig{fig:UI}.
The upper UI panel displays the color and thermal live footage from the \acs{UAV} whereas the central panel contains the reconstructed environment map in 3D including localized heat sources and the pose of the fire monitor.
The last panel contains various buttons to control the system, switch from manual to autonomous mode or reset it.
In manual mode, an operator points the fire monitor nozzle using the left joystick.
With enabled autonomy, the operator selects the heat source and authorizes the ground control station for extinguishing.
At any time, the operator easily reclaims control, \eg for safety or manual operation.

\begin{figure}[htb]
    \centering
    \resizebox{\linewidth}{!}{
    \begin{tikzpicture}
    \node[inner sep=0,anchor=north west] (image_t) at (0,0) {\includegraphics[trim=175 260 300 80,clip,height=4cm]
    {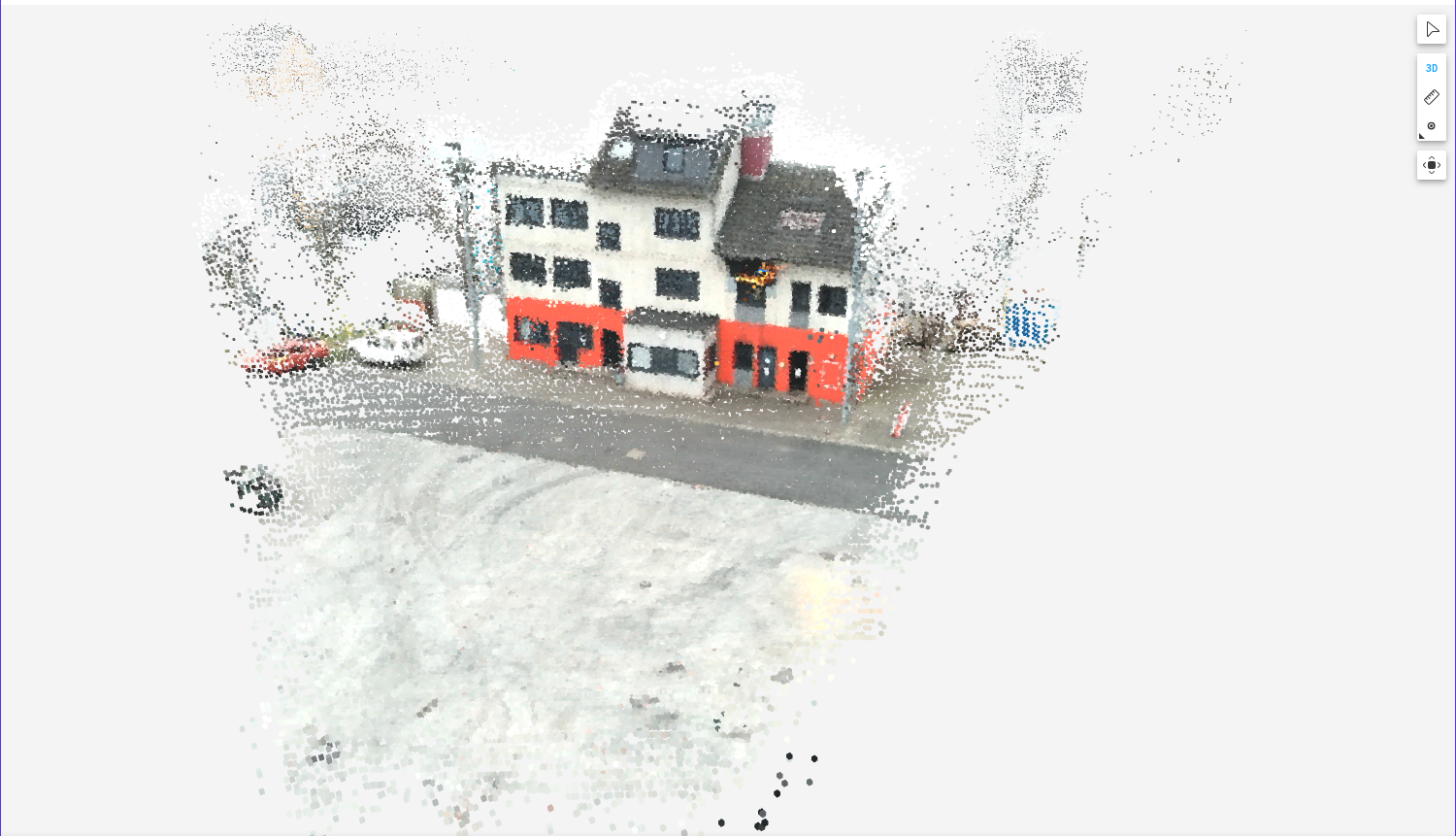}};
	\node[inner sep=0,anchor=north west,yshift=-0.1cm] (image_bl) at (image_t.south west){\includegraphics[trim=500 325 400 150,clip,height=3cm]
	{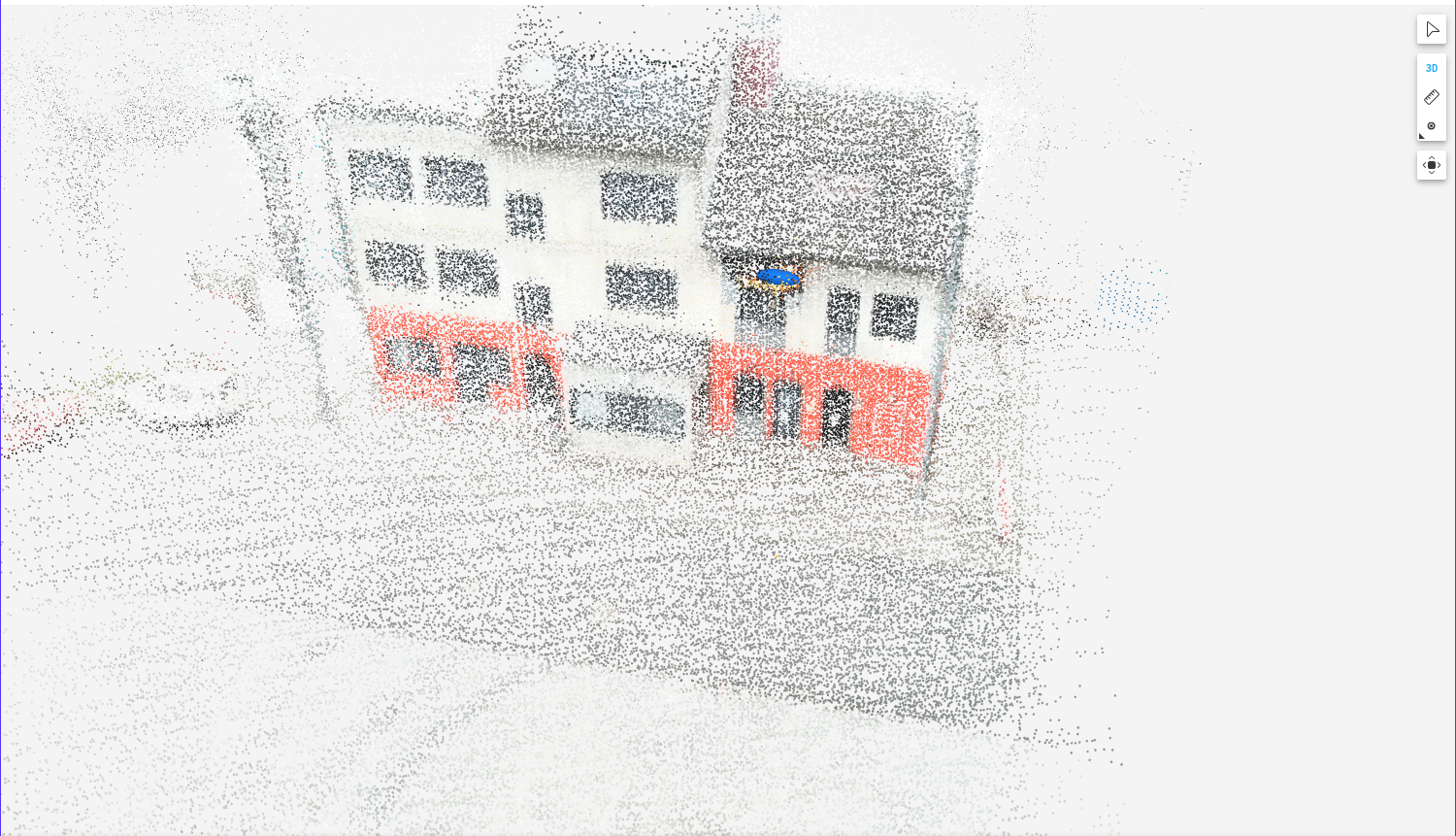}};
    \node[inner sep=0,anchor=north east,yshift=-0.1cm] (image_br) at (image_t.south east){\includegraphics[trim=475 100 400 175,clip,angle=90,height=3cm]
    {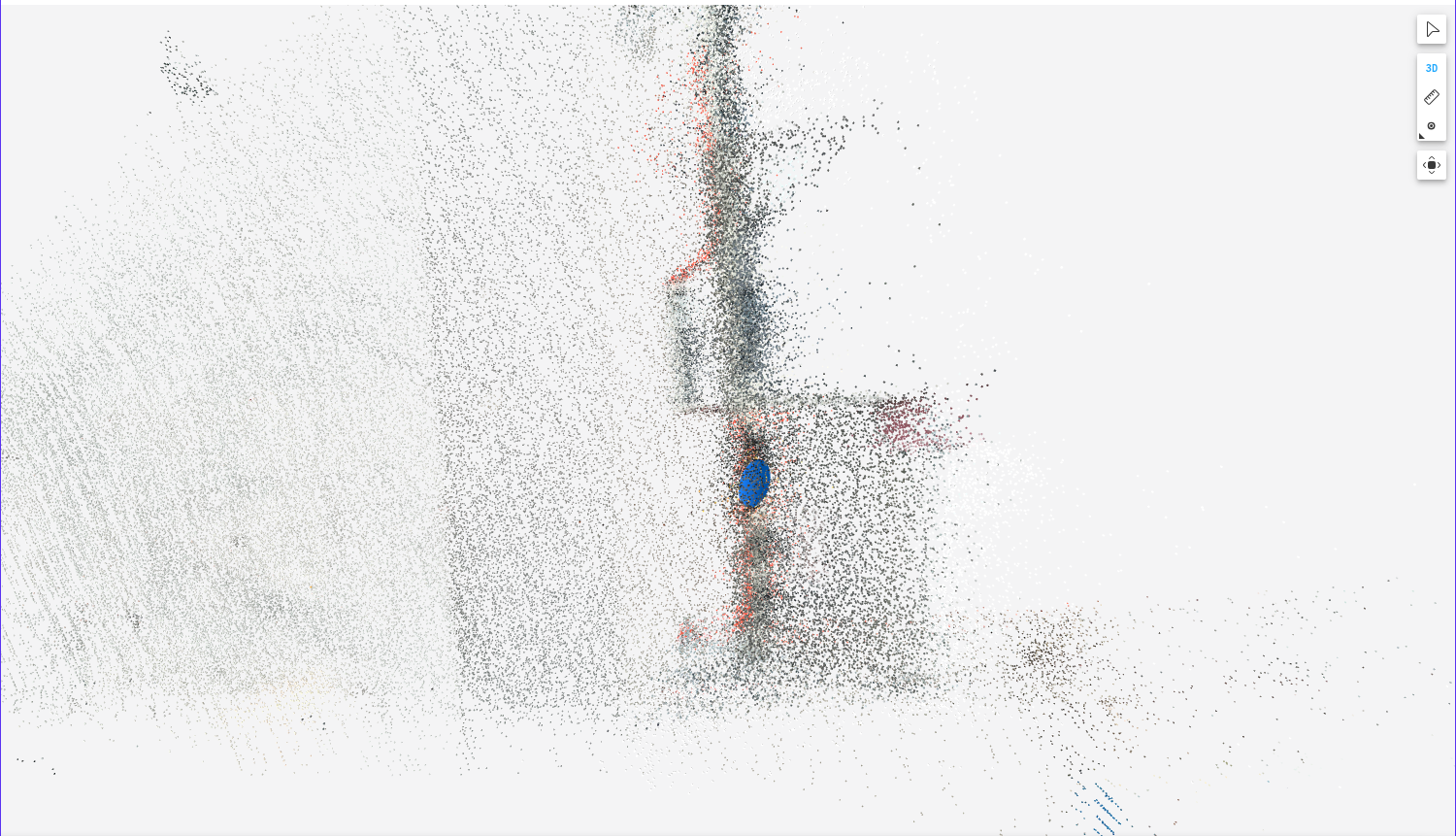}};
    \end{tikzpicture}
    }
    \caption{Localized fire (blue) at a firefighter training facility.}
    \label{fig:brandhaus}
\end{figure}

\section{Experiments}
For our experiments, the \ac{GCS} steers a DJI M3T via the DJI MSDK on the remote controller.
\ac{UAV} mission control, mapping, detection and localization run directly on the \ac{GCS} laptop with a NVIDIA GeForce RTX 5080 mobile GPU.

In initial experiments, our heat source target is a hotplate with a boiling pot of water on top.
During four autonomous flights, the heat source stays \SI{81}{\percent} of the time in the central image region with successful detections in \SI{95.8}{\percent} of the images.
Some of the missed detections are attributed to the thermal cameras' details enhancement.
The \ac{UAV} flies between the triangulation poses with up to \SI{1}{\metre\per\second} and stays in place for \SI{5}{\second}.
During these flights, the M3T alternated autonomously 22 times between the poses.

A second experiment was conducted at a firefighting training facility to test the localization with actual fire. As shown in \reffig{fig:brandhaus}, the fire is correctly localized on the building facade.

\begin{figure}[htb]
    \centering
    \resizebox{\linewidth}{!}{
    \begin{tikzpicture}
	\node[inner sep=0,anchor=north west] (image0) at (0,0)
    {\includegraphics[trim=15px 105px 15px 105px,clip,width=5cm]
    {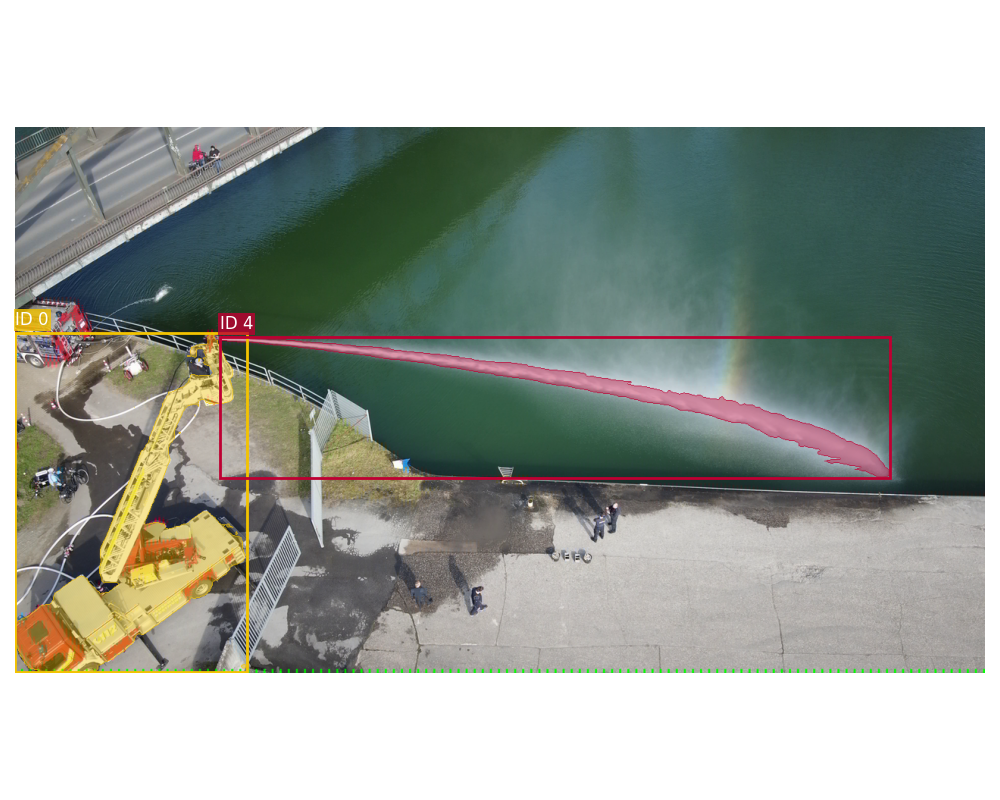}};
    \node[inner sep=0,anchor=north west,yshift=0.1cm] (image1) at (image0.south west)
    {\includegraphics[trim=15px 105px 15px 105px,clip,width=5cm]
    {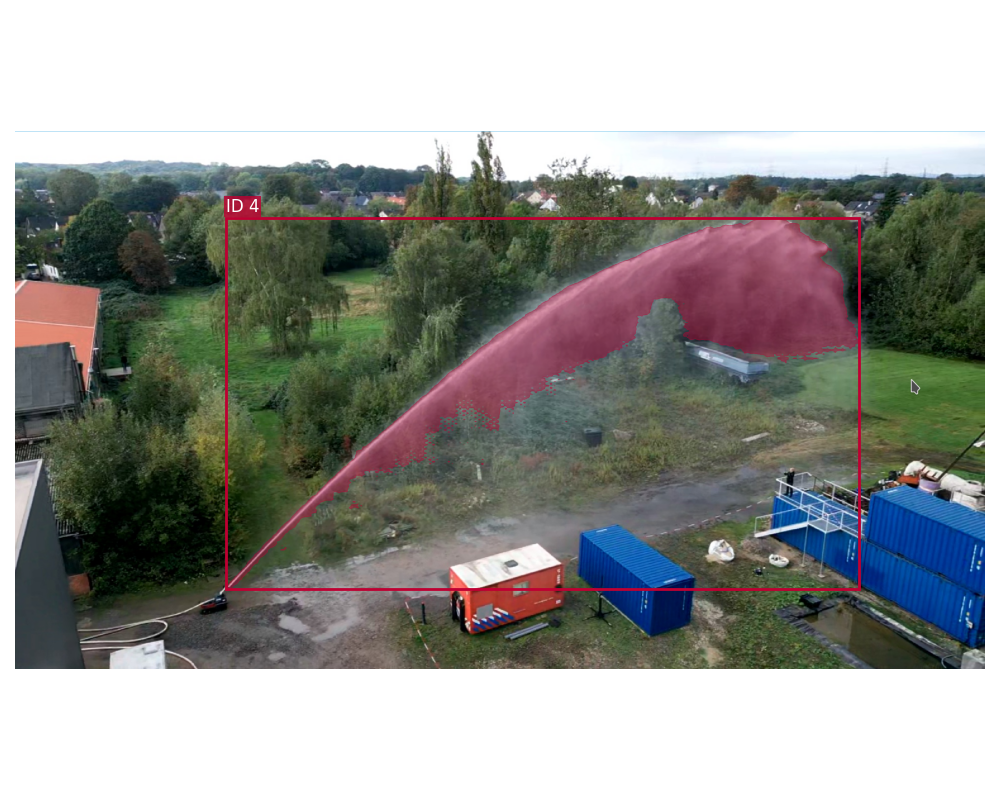}};
    \end{tikzpicture}
    }
    \caption{Detection of the water jet (red). The jet transforms from a coherent jet close to the fire monitor into spray at its end.}
    \label{fig:waterjet_detection}
\end{figure}

\begin{table}
    \renewcommand{\arraystretch}{1.2}  
    \centering
    \caption{Angular deviation of the fire monitor's motion system }
    \resizebox{\linewidth}{!}{
    \begin{tabular}{c|cc|cc|cc} 
     Speed& \multicolumn{2}{c}{\SI{10}{\percent}} & \multicolumn{2}{c}{\SI{15}{\percent}} & \multicolumn{2}{c}{\SI{20}{\percent}} \\ \hline
           & Avg. [\si{\degree}] & Std. [\si{\degree}] &  Avg. [\si{\degree}] & Std. [\si{\degree}] & Avg. [\si{\degree}] & Std. [\si{\degree}] \\ \hline
     Pitch & 0.13 & 0.077 & 0.06 & 0.271 & 0.45 & 0.252\\ \hline
     Yaw   & 0.4 & 0.24 & 0.84 & 0.299 & 1.04 & 0.275 \\
    \end{tabular}
    }
    \label{tab:angDev}
\end{table}

We tested the monitor's accuracy by alternating five times between various target angles with \SI{10}{\percent}, \SI{15}{\percent} and \SI{20}{\percent} of its maximum speed.
On average, the angular error is below \SI{0.45}{\degree} with a std. dev. of \SI{0.252}{\degree}, as shown in \reftab{tab:angDev}.
This results in a worst-case deviation of 0.46, 0.97 and \SI{1.2}{\metre}, respectively, for a ballistic trajectory at \SI{45}{\degree}.
Higher velocities resulted in significant overshooting, thus requiring a more sophisticated control loop.

Finally, we attached the fire monitor to a turntable ladder to spray water and test our system, as shown in \reffig{fig:waterjet_detection}.
Our approach successfully segmented the core of the water jet, without over segmenting the spray around it or the canal water behind.
We obtain similar results for a fire monitor mounted on a \ac{UGV}.

\begin{figure*}
    \centering
    \resizebox{\linewidth}{!}{\input{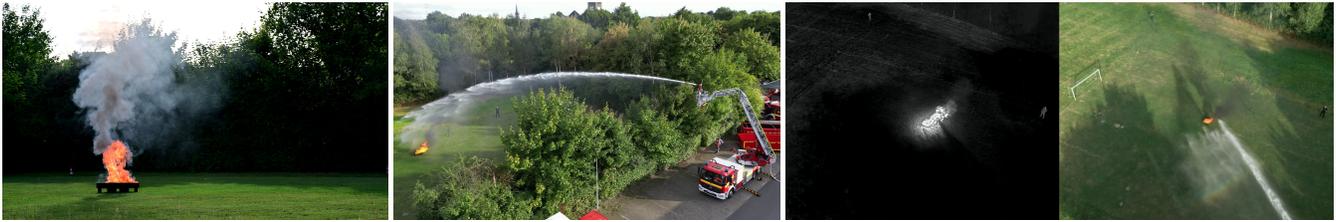}}
    \caption{Integrated system test at a fire station in Dortmund, Germany. 
    The difficult-to-reach fire (left) is hidden behind a tree line, obstructing the view towards the fire from the monitor mounted on a turntable ladder truck. The fire is clearly visible in the \acs{UAV} footage (right) and enables targeted extinguishing.}
    \label{fig:integrated_tests}
\end{figure*}

\begin{figure}[htb]
	\centering
	\resizebox{\linewidth}{!}{\input{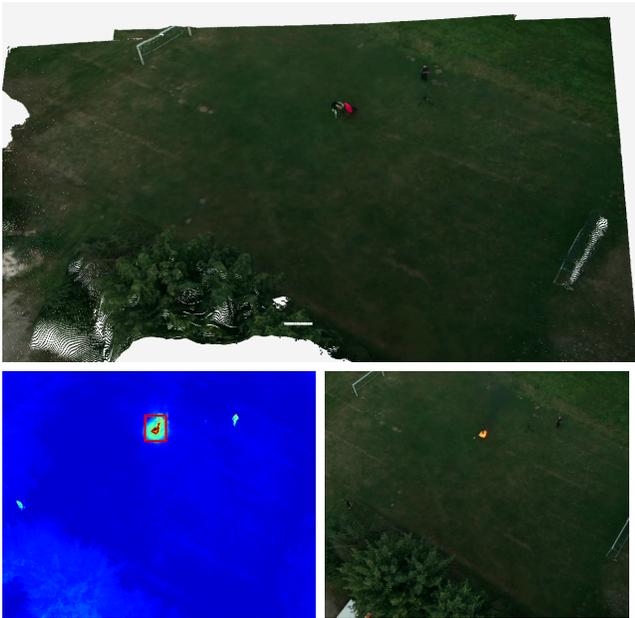}}
	\caption{Triangulated keyframes with \SI{5}{\metre} parallax enable us to localize the heat source (red ellipse) after its detection in the thermal image (bottom left). Live image (bottom right) for comparison.}
	\label{fig:exp_fire_loc}
\end{figure}

Finally, we performed a series of integrated tests with the \ac{UAV} and our fire monitor attached to a turntable ladder truck.
The overall setup is depicted in \reffig{fig:integrated_tests}.
A fire tray was placed on a field surrounded by trees, directly obstructing the view and access to the fire. As a consequence, the ladder and water trucks had to be positioned outside the tree line.
In total, we performed three test flights with the UAV. Two of those included the fire monitor spraying water.
The UAV alternates \num{17} times with a distance of \SI{5}{\metre} at a distance or around \SI{41.8}{\metre} with the fire burning around \SI{65}{\percent} of the time. 
The localized fire position was quite stable with a deviation of \SI{0.435}{\metre} for a pair of keyframes.
In between triangulations, the error increases to around \SI{1.374}{\metre}, likely due to inaccuracies in the \acs{GNSS} poses and imperfections in the synchronization between images and poses.
An example detection and localization is shown in \reffig{fig:exp_fire_loc}.

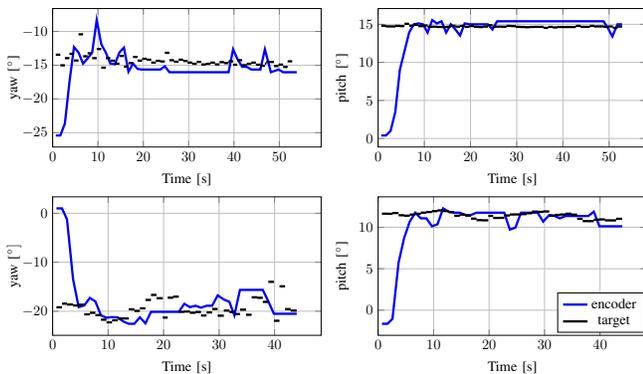
\begin{figure}[htb]
	\centering
	\resizebox{\linewidth}{!}{\begin{tikzpicture}
\begin{groupplot}[
group style={
group name=my plots,
group size=2 by 2,
vertical sep=3.5em,
horizontal sep=4.0em,
},
width=0.95\columnwidth,
height=5cm,
xlabel={Time [s]},
grid=both,
xmin=0,
cycle list name=color,
every axis plot/.append style={no markers, ultra thick}
]

\nextgroupplot[ylabel={yaw [${}^\circ$]}, ylabel style={yshift=-0.8em}, xlabel={Time [s]}]
\addplot table[x index=0, y index=1, col sep=space, each nth point=1] {control_data_rot1.dat};
\addplot[jump mark left, black] table[x index=0, y index=2, col sep=space] {control_data_target1.dat};

\nextgroupplot[ylabel={pitch [${}^\circ$]}, ylabel style={yshift=-1.0em}, xlabel={Time [s]}]
\addplot table[x index=0, y index=1, col sep=space, each nth point=1] {control_data_elev1.dat};
\addplot[jump mark left, black] table[x index=0, y index=1, col sep=space] {control_data_target1.dat};

\nextgroupplot[ylabel={yaw [${}^\circ]$}, ylabel style={yshift=-0.8em}, xlabel={Time [s]}]
\addplot table[x index=0, y index=1, col sep=space, each nth point=1] {control_data_rot2.dat};
\addplot[jump mark left, black] table[x index=0, y index=2, col sep=space] {control_data_target2.dat};

\nextgroupplot[ylabel={pitch [${}^\circ$]}, ylabel style={yshift=-1.0em}, xlabel={Time [s]}, legend style={at={(0.99,0.02)},anchor=south east}]
\addplot table[x index=0, y index=1, col sep=space, each nth point=1] {control_data_elev2.dat};
\addlegendentry{encoder};
\addplot[jump mark left, black] table[x index=0, y index=1, col sep=space] {control_data_target2.dat};
\addlegendentry{target};

\end{groupplot}
\end{tikzpicture}}
	\caption{
    Target angles (black) and reached encoder angle (blue) for the  fire monitor's pan- and tilt actuator's angle (yaw, resp. pitch) during two runs (one per row) of the integrated system test. Variations in the target angles (black) are due to updates of the fire's position.
    }
	\label{fig:control_test1}
\end{figure}

After an initial detection, the WMC quickly turns within five seconds towards the fire, as shown in \reffig{fig:control_test1}. The motors were configured to \SI{20}{\percent} of their maximum speed, as the water pressure impedes the joints capability to pitch.
To prevent the motors from stalling, the controller accepts a new target only if an angle deviates by \SI{0.5}{\degree}. Moreover, the motor interface only accepts the directional input at a low frequency.
As a result, the measured encoder angle slightly deviates from the target angle.

\section{Conclusion}
We presented a robotic assistance system for turntable ladder operations.
The system combines \ac{UAV}-based heat source localization, 3D environment mapping for safe autonomous flight and an automated fire monitor that aligns its water jet with the detected fire location.
All components are integrated into a ROS-based infrastructure that enables the data exchange between the \ac{UAV}, the fire monitor and the \ac{GCS}.
A handheld interface allows configuration and supervision of the system by the operator.
Preliminary experiments confirmed the system's ability to localize a heat source, calculate a target pose, and aim the water jet towards the chosen target pose.
Our proposed systems enables safer and more efficient firefighting in structurally complex and visually obstructed environments.

\balance
In the future, many extensions to our solution are possible, \eg the integration of a more sophisticated water jet model. 
The detected water jet could be used to optimize the jet model.
With these, direct feedback control of the jet becomes possible to further improve extinguishing accuracy.
More refined strategies for water application could lead to reduced damage by taking intact or previously burned areas into consideration.

\if\WithAuthorInfo1
\section*{Acknowledgment}
We would like to express our gratitude to the fire brigade in Dortmund, Germany, for their continued support during our experiments.
This work has been supported by the German Federal Ministry of Research, Technology and Space (BMFTR) in the projects ``Kompetenzzentrum: Etablierung des Deutschen Rettungsrobotik-Zentrums (E-DRZ)'', grant 13N16477, and ``UMDenken: Supportive monitoring of turntable ladder operations for firefighting using IR images'', grant 13N16811.
\fi

\section*{References}
\printbibliography[heading=none]

\end{document}